\documentclass{article}
\usepackage{ijcai25}

\usepackage[table]{xcolor}         
\usepackage{graphicx}
\usepackage{algorithm}
\usepackage{algorithmic}
\usepackage{amsmath,amsthm}    
\usepackage{amsmath}
\usepackage{amsfonts}
\usepackage{float}

\usepackage{caption}
\DeclareCaptionFont{ninept}{\fontsize{9pt}{11pt}\selectfont}
\usepackage{subfigure}
\usepackage[figuresright]{rotating}
\usepackage{pifont}
\usepackage{multirow}
\usepackage{subcaption}
\usepackage{stfloats}

\usepackage{times}
\usepackage{soul}
\usepackage{url}
\usepackage[hidelinks]{hyperref}
\usepackage[utf8]{inputenc}
\usepackage{booktabs}
\usepackage{algorithm}
\usepackage{algorithmic}
\usepackage[switch]{lineno}

\title{GBGC: Efficient and Adaptive Graph Coarsening via Granular-ball Computing}

\author{
    Shuyin Xia$^{1}$, 
    Guan Wang$^{1}$,
    Gaojie Xu$^{1}$,
    Sen Zhao$^{2}${\thanks{Corresponding Author}}
    and Guoyin Wang$^{3}$
    \affiliations
    $^{1}$ Chongqing Key Laboratory of Computational Intelligence, Key Laboratory of Big Data Intelligent Computing, Key Laboratory of Cyberspace Big Data Intelligent Security, Ministry of Education, 
    School of Computer Science and Technology, 
    Chongqing University of Posts and Telecommunications, Chongqing, China\\
    $^{2}$ Chongqing Key Laboratory of Computational Intelligence, Key Laboratory of Big Data Intelligent Computing, Chongqing University of Posts and Telecommunications, Chongqing, China\\
    $^{3}$ National Center for Applied Mathematics in Chongqing, Chongqing Normal University, Chongqing 401331, China
    \emails
    xiasy@cqupt.edu.cn, wangguan66@foxmail.com, s230201132@stu.cqupt.edu.cn, zhaosen@cqupt.edu.cn, wanggy@cqnu.edu.cn }

\begin{document}

\maketitle

\begin{abstract} 
The objective of graph coarsening is to generate smaller, more manageable graphs while preserving key information of the original graph. Previous work were mainly based on the perspective of spectrum-preserving, using some predefined coarsening rules to make the eigenvalues of the Laplacian matrix of the original graph and the coarsened graph match as much as possible. However, they largely overlooked the fact that the original graph is composed of subregions at different levels of granularity, where highly connected and similar nodes should be more inclined to be aggregated together as nodes in the coarsened graph. By combining the multi-granularity characteristics of the graph structure, we can generate coarsened graph at the optimal granularity. To this end, inspired by the application of granular-ball computing in multi-granularity, we propose a new multi-granularity, efficient, and adaptive coarsening method via granular-ball (GBGC), which significantly improves the coarsening results and efficiency. Specifically, GBGC introduces an adaptive granular-ball graph refinement mechanism, which adaptively splits the original graph from coarse to fine into granular-balls of different sizes and optimal granularity, and constructs the coarsened graph using these granular-balls as supernodes. In addition, compared with other state-of-the-art graph coarsening methods, the processing speed of this method can be increased by tens to hundreds of times and has lower time complexity. The accuracy of GBGC is almost always higher than that of the original graph due to the good robustness and generalization of the granular-ball computing, so it has the potential to become a standard graph data preprocessing method.
\end{abstract}

\section{Introduction}
Graphs are crucial in various fields \cite{hamilton2017representation,song2023xgcn,xv2023commerce,fout2017protein,hamaguchi2017knowledge}. Their growing size offers more information but requires substantial computational resources \cite{KumarSharmaSaxenaKumar,xu2018powerful,duan2022comprehensive}. Efficient processing methods, such as graph coarsening, are essential for processing large-scale graphs \cite{hashemi2024comprehensive}.

\begin{figure}[t]
    \centering
    \includegraphics[width=1\linewidth]{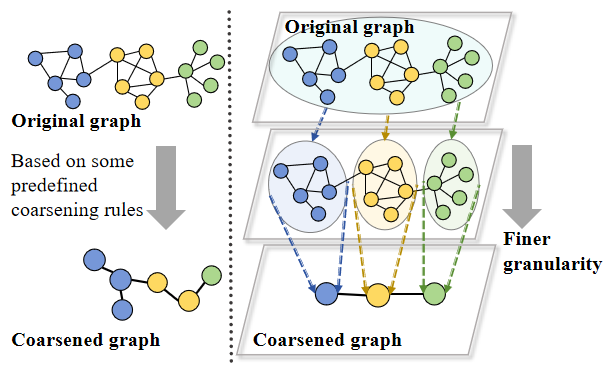}
    \caption{The differences between traditional graph coarsening methods and our proposed method.}
    \label{toyfig}
\end{figure}

Graph coarsening aims to simplify graphs while preserving their key properties and improving the efficiency of downstream tasks \cite{hashemi2024comprehensive}. Most existing works have focused on spectrum-preserving \cite{loukas2019graph,jin2020graph,ChenYaoYangChen2023}. This involves ensuring that the eigenvalues of the original graph closely match those of the coarsened graph. Although spectrum-preserving methods have many advantages, they involve the eigenvalue problem of the Laplacian matrix, which overly relies on the global properties of the graph and may hinder their application in processing large graph data due to high computational costs. Graph coarsening also finds applications in deep learning \cite{HuangZhangXiLiuZhou2021,pang2021graph}, with its challenges such as excessive information loss when coarsening is not executed properly, potentially resulting in a decline in model performance. Some works suggest that training Graph Neural Networks (GNNs) is often computationally expensive. A potential solution to alleviate the mentioned problems is dataset condensation or distillation \cite{jin2021graph,jin2022condensing}. The goal is to construct a small synthetic set to train the GNN, thereby reducing costs to some extent. However, these studies overlook the granularity characteristics of the graph structure. 

The process of graph coarsening does not eliminate any of the original nodes, but instead merges the nodes into supernodes according to some coarsening rules, and aggregates the original edges into superedges. Taking the right half of Figure \ref{toyfig} as an example, the graph typically consists of subregions with different levels of granularity, where nodes within the same subregion show a high degree of connectivity and similarity, and these nodes should be aggregated into one supernode. Initially, the original graph can be viewed as a coarsest structure, and then further refine through finer granularity until the final coarsening results are obtained. Ignoring the multi-granularity characteristics of the graph structure, as shown in the left half of Figure \ref{toyfig}, can affect the quality of the coarsened graph and the efficiency of subsequent tasks. By taking advantage of the multi-granularity characteristics of graph structure, the various structural information of the graph can be more fully understood, which can lead to more precise generation of supernodes while improving efficiency.

Actually, it is not easy to coarsen the original graph and model granularity characteristics of the graph structure, which mainly faces two challenges: \textit{\textbf{(1)} How to efficiently and effectively aggregate the nodes in the original graph into the supernodes in the coarsened graph?} Nodes with a high degree of connectivity and similarity should be more inclined to be aggregated into a supernode, which is a key consideration in graph coarsening. Inspired by the advantages of granular-ball computing in distinguishing multi-granularity of discrete data \cite{xia2023granular}, and since granular-ball computing has not yet been applied in the field of graphs, we further explore its potential in graph coarsening and propose an adaptive granular-ball graph refinement mechanism. \textit{\textbf{(2)} How to generate coarsening results according to the multi-granularity characteristics of the graph structure?} The graph structure has multiple levels of granularity features, which exhibit unique topological structures at different levels. Using this feature to guide the process of graph coarsening not only improves the efficiency of graph coarsening, but also enhances the model's ability to understand complex graph structure. Therefore, it is very important to develop a coarsening method that can capture the relationship between the nodes in the original graph and the supernodes in the coarsening graph from a multi-granularity perspective.

\footnotetext[1]{Code is available from https://anonymous.4open.science/r/GBGC. Supplementary materials are available from \url{https://github.com/Wangwangguanguan/Supplementary-materials.git.}}

To this end, we propose a new multi-granularity, efficient, and adaptive coarsening method via granular-ball. Specifically, \textbf{for challenge (1)}, the original graph is initialized as coarse-grained granular-balls, which are then iteratively split by a fine-grained binary splitting mechanism. In the initial stage, the method only uses global calculation, and the subsequent steps rely on local calculation for iterative splitting, and the computational complexity is low. \textbf{For challenge (2)}, according to the multi-granularity characteristics of the graph structure, our proposed method can adaptively use multiple granular-balls of different sizes to generate coarsening results. Extensive experiments conducted on several graph datasets show that our proposed method can outperform the state-of-the-art methods. Our contributions are summarized as follows:

\begin{itemize}
  \item We emphasize the importance of multi-granularity characteristics of the graph structure in graph coarsening.
  
  \item We propose a new graph coarsening method called \textbf{\underline{g}}ranular-\textbf{\underline{b}}all \textbf{\underline{g}}raph \textbf{\underline{c}}oarsening (GBGC), which adaptively uses different sizes of granular-balls to represent supernodes in the coarsened graph. GBGC involves global calculation (calculating the distance between all pairs of nodes in a graph) only in the initialization, thus reducing the time complexity. 
  
  \item Extensive testing and comparison of datasets confirms the validity and accuracy of GBGC, and we provide a new perspective on graph coarsening methods. 
\end{itemize}








\section{Related Work}
The spectrum of the Laplacian matrix reflects the graph's structure, and spectral distance serves as a measure of structural similarity between graphs \cite{wilson2008study}. Loukas et al. introduced ``restricted spectral similarity" (RSS), where the eigenvalues and eigenvectors of the coarsened graph approximate those of the original graph within the main feature subspace \cite{loukas2018spectrally,loukas2019graph}. Jin et al. used spectral distance to evaluate spectral preservation and developed coarsening algorithms to minimize this distance \cite{jin2020graph}. Imre et al. proposed a near-linear time-preserving sparse algorithm for large graphs \cite{imre2020spectrum}. Chen et al. applied Gromov-Wasserstein (GW) distance to limit the spectral difference between graphs and their coarsenings, proving an upper bound related to spectral difference \cite{ChenYaoYangChen2023}.

Graph coarsening methods have gained popularity in deep learning in recent years. Xie et al. proposed MLC-GCN, a multilevel coarsening classification method for GCNs that retains both local and global graph information and learns representations at multiple levels \cite{xie2020graph}. Fahrbach et al. introduced an efficient coarsening method based on the Schur complement, which reduces computational time by embedding related vertices on the coarsened graph \cite{fahrbach2020faster}. Huang et al. used graph coarsening for scalable GNN training, reducing the number of graph nodes by up to 10 times without significant accuracy loss \cite{HuangZhangXiLiuZhou2021}. Pang et al. proposed InfomaxPooling (CGIPool), which maximizes mutual information between each pooling layer’s input and the coarsened graph to preserve graph-level dependencies \cite{pang2021graph}.

Training GNNs is computationally expensive, and dataset condensation \cite{fang2024exgc} or distillation \cite{wang2018dataset} can help reduce costs by creating smaller synthetic graph datasets for training. Zhao et al. proposed a condensation method called DC, which matches network parameter gradients between a small synthetic and a large real dataset \cite{zhao2020dataset}. Jin et al. introduced GCOND, a graph compression method that reduces nodes by compressing both node features and structure in a supervised way \cite{jin2021graph}. They also proposed DosCond, a framework that uses gradient matching to effectively compress real graphs into smaller, discrete information graphs \cite{jin2022condensing}.

Above methods have significantly enhanced the efficiency of various downstream tasks, making them a focal point of ongoing research and development. Despite these advancements, many of the existing approaches tend to overlook the inherent granularity of the graph structure. This oversight can limit their effectiveness in capturing the complex relationships between nodes. In contrast, our work addresses this gap by explicitly considering the granularity characteristics of the graph. We propose a novel approach that captures and utilizes the relationships between nodes from a multi-granularity perspective, enabling more effective graph coarsening. 
 

\section{Methodology}
\subsection{Notations and problem statement}
We denote an original graph as $\mathcal{G=(V,E)}$, whose node set is $\mathcal{V}$ with size $\lvert \mathcal{V} \rvert=N$, and whose edge set is $\mathcal{E}$ with size $\lvert \mathcal{E} \rvert=E$. The adjacency matrix of $\mathcal{G}$ is denoted as $\mathbf{A} \in \{{0,1}\}^{N \times N}$, and $\mathbf{A}_{ij}=1$ if there is an edge between nodes $v_i$ and $v_j$. The Laplacian matrix is denoted as $\mathbf{L}=\mathbf{D}-\mathbf{A}$, where $\mathbf{D}$ is the degree matrix with $\mathbf{D}_{ii}=\sum_j\mathbf{A}_{ij}$. The normalized Laplacian matrix is $\hat{\mathbf{L}}=\mathbf{D}^{-\frac{1}{2}}\mathbf{L}\mathbf{D}^{-\frac{1}{2}}$. Given an original graph $\mathcal{G}$, graph coarsening methods aim to find a smaller coarsened graph denoted as $\mathcal{\overline{G}=(\mathcal{\overline{V}},\mathcal{\overline{E}}})$ to approximate $\mathcal{G}$. The number of nodes in $\mathcal{\overline{G}}$ is denoted as $\overline{N}$, the number of edges is denoted as $\overline{E}$, and $\overline{N} < N$, $\overline{E} < E$. The adjacency matrix of $\overline{\mathcal{G}}$ is denoted as $\mathbf{\overline{A}} \in \{{0,1}\}^{\overline{N} \times \overline{N}}$, and $\mathbf{\overline{A}}_{ij}=1$ if there is an edge between supernodes $\overline{v}_i$ and $\overline{v}_j$. In the matrix form, we use $\mathbf{C} \in \mathbb{R}^{N \times \overline{N}}$ to denote a projection matrix that maps nodes in $\mathcal{V}$ to supernodes in $\mathcal{\overline{V}}$, and $\mathbf{C}_{ij}=1$ if node $i$ in $\mathcal{V}$ is mapped to supernode $j$ in $\mathcal{\overline{V}}$. The Laplacian matrix of the coarsened graph can be denoted as $\mathbf{\overline{L}}=\mathbf{C}^T\mathbf{L}\mathbf{C}$. This computation ensures that the spectral properties of the original graph are preserved in the coarsened graph. 


Our proposed GBGC seeks to obtain a coarsened graph $\mathcal{\overline{G}}$ to approximate $\mathcal{G}$. The original graph $\mathcal{G}$ is refined into multiple non-overlapping subdomains as follows:

\begin{align}
\mathcal{G}=\{\mathcal{\widetilde{G}}_1,\mathcal{\widetilde{G}}_2,...,\mathcal{\widetilde{G}}_t\}(t=\overline{N}), \\\mathcal{\widetilde{G}}_i=(\mathcal{\widetilde{V}}_i,\mathcal{\widetilde{E}}_i),i=1,2,...,t,
\end{align}

\noindent where $\mathcal{\widetilde{V}}_i$ and $\mathcal{\widetilde{E}}_i$ represent the set of nodes and edges of the graph $\mathcal{\widetilde{G}}_i$ contained inside the granular-ball $\mathcal{GB}_i$, with $\mathcal{\widetilde{V}}_i\subset\mathcal{V}$ being a subset of nodes from $\mathcal{G}$. $\overline{N}$ is the number of supernodes in $\mathcal{\overline{G}}$, and each $\mathcal{GB}$ corresponds to a supernode in $\mathcal{\overline{G}}$. In the process of graph coarsening implementation, the information loss between the original graph and the coarsened graph should be minimized, the number of granular-balls should be as few as possible, and the adaptive quality requirements should be met. 

\subsection{Model introduction}
\begin{figure*}[tb!]
    \centering
    \includegraphics[width=1.0\linewidth]{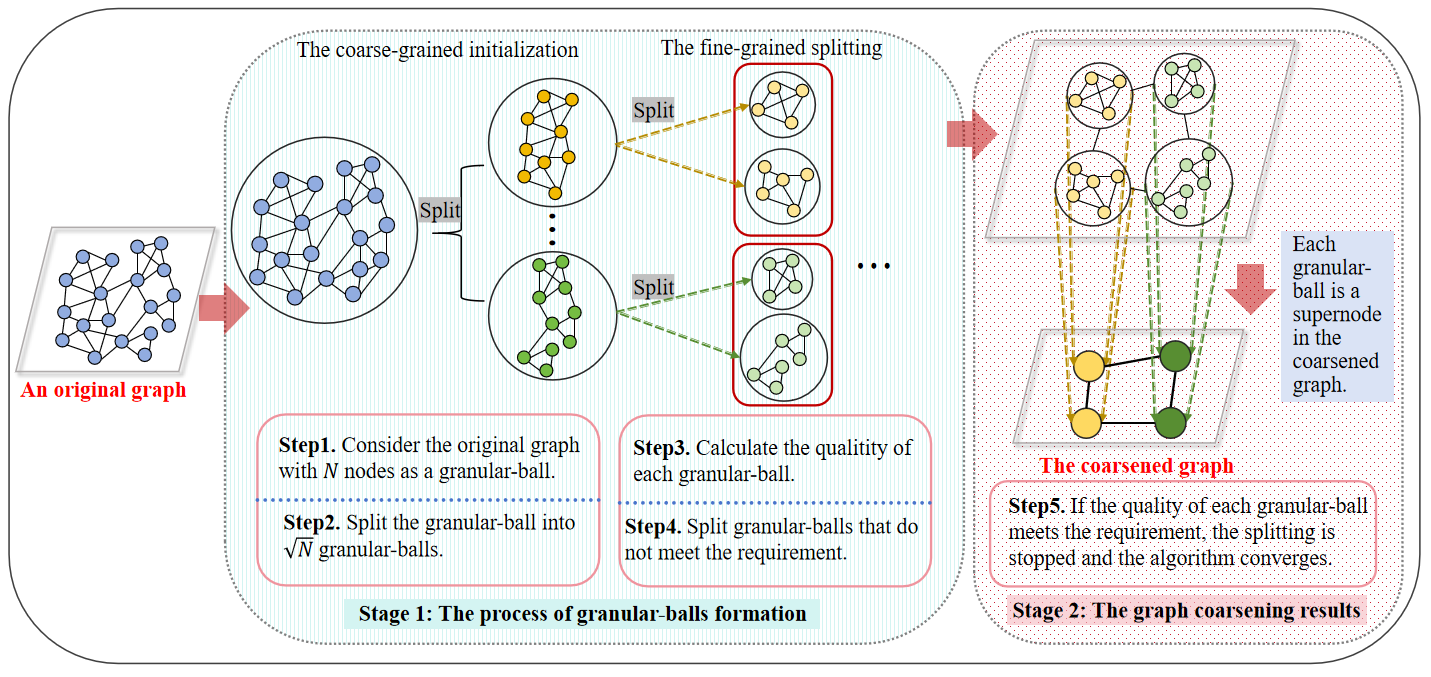}
    \caption{The framework of our method. GBGC consists of two main stages. As shown in this figure, the first stage is the process of granular-balls formation, which consists of four steps. Specifically, Steps 1 and 2 depict the initialization of coarse-grained granular-balls, while Steps 3 and 4 demonstrate the fine-grained binary splitting process. The second stage consists of Step 5, where GBGC generates the final coarsening results.}
    \label{framework}
\end{figure*}

Figure \ref{framework} illustrates the overall framework of GBGC, which consists of two stages: (I) \textit{the process of granular-balls formation}; (II) \textit{the graph coarsening results}.

\subsubsection{(I) The process of granular-balls formation}
This stage consists of four steps. \textbf{Step 1} is based on the law of ``global precedence'' \cite{Chen1982}, the original graph as the coarsest granular-ball. \textbf{Step 2} generates $\sqrt{N}$ initial granular-balls. Mathematically, 

\begin{equation}
\mathcal{GB}_{init}=\{\mathcal{GB}_1,\mathcal{GB}_2,...,\mathcal{GB}_s\}(s=\sqrt{N}). 
\end{equation}

The value of $\sqrt{N}$ is an empirical choice, referenced from previous work \cite{xie2020new,yu2001upper}, which has been shown to balance the trade-off between size and quantity of granular-balls. 

The procedure for generating $\mathcal{GB}_{init}$ is as follows. Firstly, we select the node with the highest degree $v_{center}=\arg\max\limits_{v\in\mathcal{V}}deg(v)$ as the initial center. Subsequently, breadth-first search (BFS) starts with the $v_{center}$, traversing layer by layer until the number of nodes at layer $i$ exceeds $\sqrt{N}$:

\begin{equation}
\lvert \mathcal{N}_i \rvert>\sqrt{N},
\end{equation}

\noindent where $\lvert \mathcal{N}_i \rvert$ denotes the number of nodes at layer $i$. Then delete these nodes from $\mathcal{V}$:

\begin{equation}
\mathcal{V}'=\mathcal{V}\setminus\mathcal{N}_i.
\end{equation}

Continue to find the next center with the highest degree $v'_{center}=\arg\max\limits_{v\in\mathcal{V}'}deg(v)$ in $\mathcal{V}'$, and so on through the entire graph to determine $\mathcal{GB}_{init}$. 

All nodes participate in the calculation, so global calculation is involved in this step. This approach not only accelerates subsequent granular-ball splitting processes but also provides an effective starting point for granular-ball coarsening in graphs with a small number of nodes. \textbf{Step 3} is to calculate the quality of each granular-ball, which is defined as follows.




\textbf{Definition 1} For a granular-ball $\mathcal{GB}$, its quality is formulated as follows: 

\begin{equation}
quality(\mathcal{GB})=\frac{\widetilde{E}}{\widetilde{N}} + \frac{\sum_{\widetilde{i}\neq \widetilde{j}\neq \widetilde{k}}\mathbf{\widetilde{A}}_{\widetilde{i}\widetilde{j}}\mathbf{\widetilde{A}}_{\widetilde{j}\widetilde{k}}\mathbf{\widetilde{A}}_{\widetilde{k}\widetilde{i}}}{\sum_{\widetilde{i}\neq \widetilde{j}\neq \widetilde{k}}\mathbf{\widetilde{A}}_{\widetilde{i}\widetilde{j}}\mathbf{\widetilde{A}}_{\widetilde{i}\widetilde{k}}},
\end{equation}

\noindent where $\widetilde{E}$ and $\widetilde{N}$ represent the number of edges and nodes within the graph contained inside $\mathcal{GB}$, respectively. $\mathbf{\widetilde{A}}_{mn}$ indicates the presence ($\mathbf{\widetilde{A}}_{mn}=1$) or absence ($\mathbf{\widetilde{A}}_{mn}=0$) of an edge between nodes $\widetilde{m}$ and $\widetilde{n}$ ($\widetilde{m}$, $\widetilde{n}=\widetilde{i},\widetilde{j},\widetilde{k}$) inside $\mathcal{GB}$.

Quality is composed of two components: the average degree and the global clustering coefficient (also known as the transitivity). The first component quantifies the average number of edges connected to each node in the graph, serving as an intuitive measure of the connectivity within the graph. In the second component, $\sum_{\widetilde{i}\neq \widetilde{j}\neq \widetilde{k}}\mathbf{\widetilde{A}}_{\widetilde{i}\widetilde{j}}\mathbf{\widetilde{A}}_{\widetilde{j}\widetilde{k}}\mathbf{\widetilde{A}}_{\widetilde{k}\widetilde{i}}$, represents the number of actual triangles present in the graph. In contrast, $\sum_{\widetilde{i}\neq \widetilde{j}\neq \widetilde{k}}\mathbf{\widetilde{A}}_{\widetilde{i}\widetilde{j}}\mathbf{\widetilde{A}}_{\widetilde{i}\widetilde{k}}$ denotes the total number of possible triples in the graph, where a triple is defined as a central node connected to two other nodes, regardless of whether the two peripheral nodes are directly connected. The second component measures the local clustering of nodes within the graph, and its computation relies on the number of triangles and triples.

The higher the quality of the granular-ball, the more tightly connected the nodes inside the granular-ball. This method of quantifying the quality of the granular-ball ensures that the structural properties of the graph are considered and balanced in subsequent splitting of the granular-ball. 


\textbf{Step 4} is to perform a fine-grained binary splitting for each granular-ball $\mathcal{GB}_j$ in $\mathcal{GB}_{init}$. The fine-grained splitting identifies the two nodes with the highest degree of parent granular-ball ($\mathcal{GB}_j$) as the splitting centers of the two child granular-balls ($\mathcal{GB}_{j_{A}}$ and $\mathcal{GB}_{j_{B}}$), which are denoted as 
\begin{equation}
v_A=\arg\max\limits_{v\in\mathcal{\widetilde{V}}_{j}}deg(v), v_B=\arg\max\limits_{v\in\mathcal{\widetilde{V}}_{j}\setminus{v_A}}deg(v), 
\end{equation}

\noindent where $\mathcal{\widetilde{V}}_{j}$ represents the set of nodes within the granular-ball $\mathcal{GB}_j$. Each center node traverses the remaining nodes layer by layer through BFS, and other non-center nodes are assigned to the node according to who was first traversed by a center node’s BFS. Only the nodes within granular-balls are involved in the calculation, so Steps 3 and 4 involve local calculation. The adaptive splitting condition of parent granular-ball splitting is given by the following formula:

\begin{equation}
{quality}(\mathcal{GB}_{j_{A}}+\mathcal{GB}_{j_{B}}) > {quality}(\mathcal{GB}_j),
\end{equation}
\noindent where ${quality}(\mathcal{GB}_{j_{A}}+\mathcal{GB}_{j_{B}})$ denotes the sum of the qualities of the two child granular-balls, and $quality(\mathcal{GB}_j)$ denotes the quality of the parent granular-ball. This formula is employed to decide whether to proceed with splitting granular-balls. If there is a notable difference in quality between the parent granular-ball and its child granular-balls, suggesting that further splitting improves the overall system performance, the splitting operation will continue. This strategy ensures both the precision and efficiency of the splitting process, preventing unnecessary computational complexity from arising.



\subsubsection{(II) The graph coarsening results}
This stage consists of \textbf{Step 5}, which is to generate the coarsening results. In the process of graph coarsening, GBGC splits the nodes of the original graph from coarse-grained to fine-grained, and generates the supernodes of the coarsened graph. Therefore, each granular-ball $\mathcal{GB}$ corresponds to a supernode in the coarsened graph $\mathcal{\overline{G}}$. Mathematically, 

\begin{equation}
\mathcal{\overline{V}}=\{\mathcal{GB}_1,\mathcal{GB}_2,...,\mathcal{GB}_t\}(t=\overline{N}), 
\end{equation}

\noindent which covers all nodes in $\mathcal{G}$. If there is already an edge between the nodes contained within two different granular-balls, then an edge is created between the two granular-balls. The process can be represented as follows:

\begin{equation}
\begin{tiny}
\mathcal{\overline{E}}\!=\!\{(\mathcal{GB}_i,\mathcal{GB}_j)\!\mid\!\exists e\!=\!(u,v)\!\in\! \mathcal{E},u\!\in\!\mathcal{\widetilde{V}}_{i},v\!\in\!\mathcal{\widetilde{V}}_{j},\mathcal{GB}_i\!\neq\!\mathcal{GB}_j\}.
\end{tiny}
\end{equation}

This methodology enhances the representation of graph by focusing on broader connectivity patterns, while maintaining the critical properties of the graph. 

\subsection{Theoretical analysis of GBGC}
Due to the difference in the number of nodes between the original graph $\mathcal{G}$ and the coarsened graph $\mathcal{\overline{G}}$, we introduce appropriate intrinsic functionals such as the Rayleigh quotient (which is a quantity related to the structure of the graph and does not vary with the arrangement of nodes). The projection matrix $\mathbf{C}$, where $\mathbf{C}_{ij}=1$ if $v_i$ belongs to $\mathcal{GB}_j$, ensures that the spectral information of the two graphs becomes comparable \cite{cai2021graph}. Therefore, we first give the definitions of the Rayleigh quotient of the original graph and the coarsened graph respectively. Then the coarsening effect of GBGC is proved by the Rayleigh quotient.

\textbf{Definition 2} For an original graph $\mathcal{G}$, its Rayleigh quotient is defined as:
\begin{equation}
R_o(\mathbf{\mathbf{L},x})=\frac{\mathbf{x}^T\mathbf{L}\mathbf{x}}{\mathbf{x}^T\mathbf{x}},
\end{equation}

\noindent where $\mathbf{L}$ is the Laplace matrix of $\mathcal{G}$, $\mathbf{x}$ is the eigenvector of $\mathbf{L}$. According to the definition of the projection matrix $\mathbf{C}$ in Section 3.1, the Rayleigh quotient of the coarsened graph obtained by GBGC is given, which is defined as follows:

\textbf{Definition 3} For the coarsened graph $\mathcal{\overline{G}}$, its Rayleigh quotient is defined as:
\begin{equation}
R_c(\mathbf{\mathbf{\overline{L}},\mathbf{\overline{x}}})=\frac{(\mathbf{C}^T\mathbf{x})^T\mathbf{\overline{L}}(\mathbf{C}^T\mathbf{x})}{(\mathbf{C}^T\mathbf{x})^T(\mathbf{C}^T\mathbf{x})},
\end{equation}

\noindent where $\mathbf{\overline{L}}$ is the Laplace matrix of $\mathcal{\overline{G}}$. 

\textbf{Theorem 1} Assuming $\mathbf{C}$ is a generalized orthogonal matrix (i.e. $\mathbf{C}^T\mathbf{C}\approx\mathbf{I}$, where $\mathbf{I}$ is the identity matrix), we have:

\begin{equation}
R_c(\mathbf{\mathbf{\overline{L}},\mathbf{\overline{x}}})\approx R_o(\mathbf{\mathbf{L},x}).
\end{equation}

\begin{proof}
Considering the assumption that $\mathbf{C}$ is a generalized orthogonal matrix. For the denominator of $R_c(\mathbf{\mathbf{\overline{L}},\mathbf{\overline{x}}})$, using the orthogonality of $\mathbf{C}$, we get $(\mathbf{C}^T\mathbf{x})^T(\mathbf{C}^T\mathbf{x})\approx \mathbf{x}^T\mathbf{C}\mathbf{C}^T\mathbf{x}=\mathbf{x}^T\mathbf{I}\mathbf{x}=\mathbf{x}^T\mathbf{x}$. And then, for the numerator of $R_c(\mathbf{\mathbf{\overline{L}},\mathbf{\overline{x}}})$, we have $(\mathbf{C}^T\mathbf{x})^T\mathbf{\overline{L}}(\mathbf{C}^T\mathbf{x}) \approx \mathbf{x}^T\mathbf{C}\mathbf{\overline{L}}\mathbf{C}^T\mathbf{x}$. According to $\mathbf{\overline{L}}=\mathbf{C}^T\mathbf{L}\mathbf{C}$, we obtain $\mathbf{C}\mathbf{\overline{L}}\mathbf{C}^T \approx \mathbf{L}$. Thus, $\mathbf{x}^T\mathbf{C}\mathbf{\overline{L}}\mathbf{C}^T\mathbf{x}\approx\mathbf{x}^T\mathbf{L}\mathbf{x}$. Finally, using these approximations, the Rayleigh quotient of the coarsened graph obtained by GBGC can be approximated as $R_c(\mathbf{\mathbf{\overline{L}},\mathbf{\overline{x}}})\approx R_o(\mathbf{\mathbf{L},x})$.
\end{proof}

\section{Experiments}

In this section, we design experiments to evaluate our GBGC and answer the following research questions (RQs): \textbf{RQ1:} Does our method achieve better performance compared to the state-of-the-art methods? \textbf{RQ2:} How does GBGC perform with respect to time consume? \textbf{RQ3:} What impact do the key components have on GBGC's performance? \textbf{RQ4:} How do hyperparameters effect GBGC's performance? \textbf{RQ5:} How does GBGC perform in visualization?

\subsection{Experimental setup}

\textbf{Datasets:} We use several standard graph classification datasets, which are shown in Table \ref{datasetable}. See Appendix for implementation details.

\begin{table}
    \centering
        \begin{tiny}
    \begin{tabular}{ccccc}
       \toprule
Dataset  & Size & $\lvert \mathcal{V} \rvert$ & $\lvert \mathcal{E} \rvert$ \\
\midrule
MUTAG \cite{debnath1991structure}    &  188 & 17.93 & 19.79 \\
PROTEINS \cite{borgwardt2005protein} &  1113 & 39.06 & 72.82 \\
IMDB-BINARY \cite{cai2018simple}    &  1000 & 19.77 & 96.53 \\
NCI109 \cite{morris2020weisfeiler}    &  4127 & 29.68 & 32.13 \\
DHFR \cite{sutherland2003spline}    &  756 & 42.43 & 44.54 \\
BZR \cite{vincent2021online}      &  405 & 35.75 & 38.36 \\
Tox21\_AR\-LBD\-testing \cite{cooper2019improving}      &  253 & 21.85 & 22.73 \\
OVCAR-8H \cite{morris2020tudataset}     &  39253 & 46.67 & 48.70 \\
P388H \cite{morris2020tudataset}     &  40651 & 40.45 & 41.89 \\
SF-295H \cite{morris2020tudataset}     &  39030 & 46.65 & 48.68 \\
DD \cite{morris2020tudataset}     &  1178 & 284.32 & 715.66 \\
\bottomrule
    \end{tabular}
        \end{tiny}
    \caption{Summary of dataset information. $\lvert \mathcal{V} \rvert$ and $\lvert \mathcal{E} \rvert$ denote the average number of nodes and edges, respectively. All graphs are undirected in this paper.}
    \label{datasetable}
\end{table}



\textbf{Baselines:} We compare with six baseline methods also for graph-level tasks: Variation Neighborhood Graph Coarsening (VNGC) \cite{loukas2019graph}; Variation Edge Graph Coarsening (VEGC) \cite{loukas2019graph}; Multilevel Graph Coarsening (MGC) \cite{jin2020graph}; Spectral Graph Coarsening (SGC) \cite{jin2020graph}; Dataset One-Step Condensation (DosCond) \cite{jin2022condensing}; Kernel Graph Coarsening (KGC) \cite{ChenYaoYangChen2023}.

\textbf{Evaluation index of coarsening effect:} In this paper, we mainly use accuracy (ACC) and spectral distance (SD) to measure the effect of coarsening of GBGC. According to the reference \cite{wilson2008study}, since the spectrum of the Laplacian distills information about the graph structure, one may interpret the spectral distance as a proxy for the structural similarity of the two graphs. The spectral distance between the original graph and the coarsened graph is defined by calculating the Euclidean distance between the eigenvalue vectors of the two graphs, i.e., $SD(\mathcal{G},\mathcal{\overline{G}})=\sqrt{\sum_i(\lambda_i^{(\mathcal{G})}-\lambda_i^{(\mathcal{\overline{G}})})^2}$, where $\lambda_i^{(\mathcal{G})}$ and $\lambda_i^{(\mathcal{\overline{G}})}$ are the $i$-th eigenvalues of the Laplace matrix of the original and coarsened graphs, respectively. When the spectra have different sizes, then the smaller one can be padded with zeros (while maintaining the magnitude ordering). This is equivalent to adding disjoint nodes to the smaller graph to make both graphs have the same number of nodes.

\begin{table*}
    \centering
    \begin{small}
    \resizebox{\textwidth}{!}{%
    \begin{tabular}{cccccccccccc}
        \toprule
        Datasets & MUTAG & PROTEINS & IMDB-BINARY & NCI109 & DHFR  & BZR & Tox21\_AR-LBD & OVCAR-8H & P388H & SF-295H & DD \\
        \midrule
       $r_a$    & 0.35 & 0.24 & 0.19 & 0.38 & 0.35 & 0.40 & 0.34 & 0.38 & 0.37 & 0.38 & 0.20 \\
        \midrule
        VNGC    & 77.75 $\pm$ 2.15 & 63.45 $\pm$ 0.92 & 49.60 $\pm$ 0.78 & 61.68 $\pm$ 0.22 & 62.03 $\pm$ 1.66  & 77.74 $\pm$ 1.45 &  \underline{96.60 $\pm$ 0.88} & 91.44$\pm$0.07 & 91.35$\pm$0.08 & 91.79$\pm$0.06 & 65.70$\pm$1.25   \\
        VEGC    & 81.93 $\pm$ 2.24 & 63.90 $\pm$ 1.00 & 61.50 $\pm$ 1.88 & 62.03 $\pm$ 0.40 & 68.12 $\pm$ 1.04 & 78.52 $\pm$ 2.35 & 94.93 $\pm$ 1.23 & 91.88$\pm$0.08 & 91.86$\pm$0.12 &	92.09$\pm$0.11 & 69.19$\pm$1.28 \\
        MGC & 84.56 $\pm$ 1.75 & \underline{65.88 $\pm$ 1.28} & 65.70 $\pm$ 1.18 & \underline{65.09 $\pm$ 0.72} & 71.42 $\pm$ 1.63 & 76.06 $\pm$ 2.01 & 95.49 $\pm$ 1.56 & 91.80$\pm$0.11 & 91.51$\pm$0.13 & 91.93$\pm$0.07 & OOM \\
        SGC    & 83.54 $\pm$ 2.27 & 64.81 $\pm$ 1.58 & \underline{66.60 $\pm$ 1.29} & 62.95 $\pm$ 0.59 & 63.21 $\pm$ 1.83 & 78.04 $\pm$ 0.93 & 95.46 $\pm$ 0.72 & 91.45$\pm$0.14	& 91.24$\pm$0.15 & 91.59$\pm$0.06 & 67.39$\pm$1.26 \\
        DosCond    & 70.00 $\pm$ 1.68 & 63.89 $\pm$ 1.28 & 59.17 $\pm$ 0.21 & 56.75 $\pm$ 1.08 & \underline{71.66 $\pm$ 0.70} & 66.50 $\pm$ 1.29 & 87.43 $\pm$ 0.74 & 92.11$\pm$1.35	&  \textbf{92.93$\pm$1.05} & 85.49$\pm$1.03 & \textbf{70.33$\pm$1.83} \\
        KGC     & 77.60 $\pm$ 2.01 & 64.62 $\pm$ 1.32 & 63.10 $\pm$ 1.40 & 57.75 $\pm$ 0.70 & 64.28 $\pm$ 1.81 & 78.04 $\pm$ 1.96 & 94.90 $\pm$ 0.95 & 91.31$\pm$0.12 &	90.91$\pm$0.13 &	91.44$\pm$0.08 & OOM \\ 
        GBGC      &  \textbf{90.94 $\pm$ 2.12} &  65.34 $\pm$ 0.98 &  66.10 $\pm$ 1.15 &  \textbf{66.53 $\pm$ 0.72}  &  \textbf{72.62 $\pm$ 1.69} &  \textbf{80.01 $\pm$ 0.80} & \textbf{97.19 $\pm$ 1.19} &  \textbf{92.53$\pm$0.06} & \underline{92.33$\pm$0.11} &	 \textbf{92.63$\pm$0.09} &  \underline{69.44$\pm$1.65} \\
        \midrule
        FULL     & \underline{85.12 $\pm$ 2.44} & \textbf{66.61 $\pm$ 0.99} & \textbf{69.30 $\pm$ 1.34} & 64.91 $\pm$ 0.81 & 71.42 $\pm$ 1.26 & \underline{79.78 $\pm$ 2.05} & 94.31 $\pm$ 0.79 & \underline{92.35$\pm$0.14}	& 92.28$\pm$0.14 & \underline{92.54$\pm$0.08} & 66.72$\pm$1.01  \\
        \bottomrule
    \end{tabular}%
    }
    \end{small}
    \caption{Comparison of classification performance with the baseline methods, FULL indicates the performance of the dataset without coarsening. The best results are bolded, the second results are underlined, and ``OOM" means out of memory.}
    \label{acctable}
\end{table*}

\begin{table*}
    \centering
        \begin{small}
        \resizebox{\textwidth}{!}{%
    \begin{tabular}{cccccccccccc}
        \hline
        Datasets & MUTAG & PROTEINS & IMDB-BINARY & NCI109 & DHFR  & BZR & Tox21\_AR-LBD & OVCAR-8H & P388H & SF-295H & DD \\
        \hline
        VNGC    & 3.40 & 41.10 & \underline{18.31} & 109.37 & 33.66  & 15.87 & 4.00 & 2005.11 &	2065.66	& 2111.72 &  427.02  \\
        VEGC    & \underline{2.80} & \underline{25.00} & 19.85 & \underline{58.92} & \underline{14.53}  & \underline{6.95} & \underline{2.57} & \underline{946.53} &	\underline{1020.55} &	\underline{973.16} & \underline{359.26} \\
        MGC & 3.36 & 2227.51 & 38.90 & 344.16 & 138.92  & 43.07 & 6.14  & 15661.21 &	11146.01	& 15837.83 & OOM \\
        SGC    & 27.04 & 1336.21 & 41.67 & 1375.79 & 443.26  & 206.77 & 32.83 & 38465.95	& 27621.03 &	38165.46 & 163677.65 \\
        KGC     & 7.13 & 81.38 & 23.97 & 208.59 & 55.03  & 27.03 & 6.98  & 3517.12 &	2934.62 &	3256.30 & OOM \\  
        GBGC     & \textbf{0.44} & \textbf{11.70} & \textbf{3.63} & \textbf{17.74} & \textbf{9.37}  & \textbf{2.43} & \textbf{0.96}  & \textbf{350.46}	& \textbf{281.59}	& \textbf{333.02} & \textbf{44.61}\\
        \hline
    \end{tabular}}
        \end{small}
    \caption{Running time comparison table (Unit: Second). The best results are bolded, the second results are underlined, and ``OOM" means out of memory.}
    \label{timetable}
\end{table*}

\subsection{Performance of coarsening method for graph classification \textbf{(RQ1)}}
Graph classification is a key data mining task, however, the complexity and diversity of large-scale graph data bring great challenges to classification. Graph coarsening can reduce graph size, speeding up training and inference for graph classification models. However, improper coarsening may cause information loss, reducing accuracy. Thus, we evaluate various coarsening methods by comparing their performance with the same classifier applied to the original graph.

Our GBGC is free of randomness. To evaluate comparison models thoroughly, we used multiple random seeds and applied cross-validation with repeated runs. The DosCond method, which doesn’t use a ratio for computation, averages results from ten repeated experiments to ensure more reliable accuracy metrics. This reduces the impact of chance errors on final results, making them more robust. To ensure a fair comparison, the identical coarsening parameter $r_a$, as determined by the adaptive coarsening process of GBGC, is utilized across the VNGC, VEGC, MGC, SGC, and KGC. The performance results are shown in Table \ref{acctable}. From Table \ref{acctable}, we have the following observations:

\begin{itemize}
\item \textbf{It is evident that GBGC achieves higher graph classification accuracy compared to baseline methods on most datasets.} GBGC performs better on most datasets than comparison baseline methods and FULL. For the P388H dataset, the accuracy of GBGC was not significantly different from that of DosCond. However, GBGC consistently excels in graph classification, especially in maintaining stable performance across multiple datasets. This indicates that GBGC better preserves the original graph's properties while reducing its size, effectively capturing structural information and improving classification accuracy.

\item \textbf{GBGC demonstrates high accuracy in graph classification after coarsening the original graph.} GBGC outperforms the full dataset in classification accuracy on most datasets. This is due to the coarsened graph reducing noise and the granular-ball effectively capturing data distribution. Similar to other work related to granular-ball computing, GBGC is robust and generalizable. It uses multi-granularity representation and granular-ball splitting to generate supernodes, preserving key graph structure information for better classification. However, GBGC performs worse than FULL on IMDB-BINARY and PROTEINS datasets, likely because these datasets have better distribution and less noise, reducing the benefits of coarsening.
\end{itemize}

\begin{figure}[tb!]
    \centering
    \includegraphics[width=0.98\linewidth]{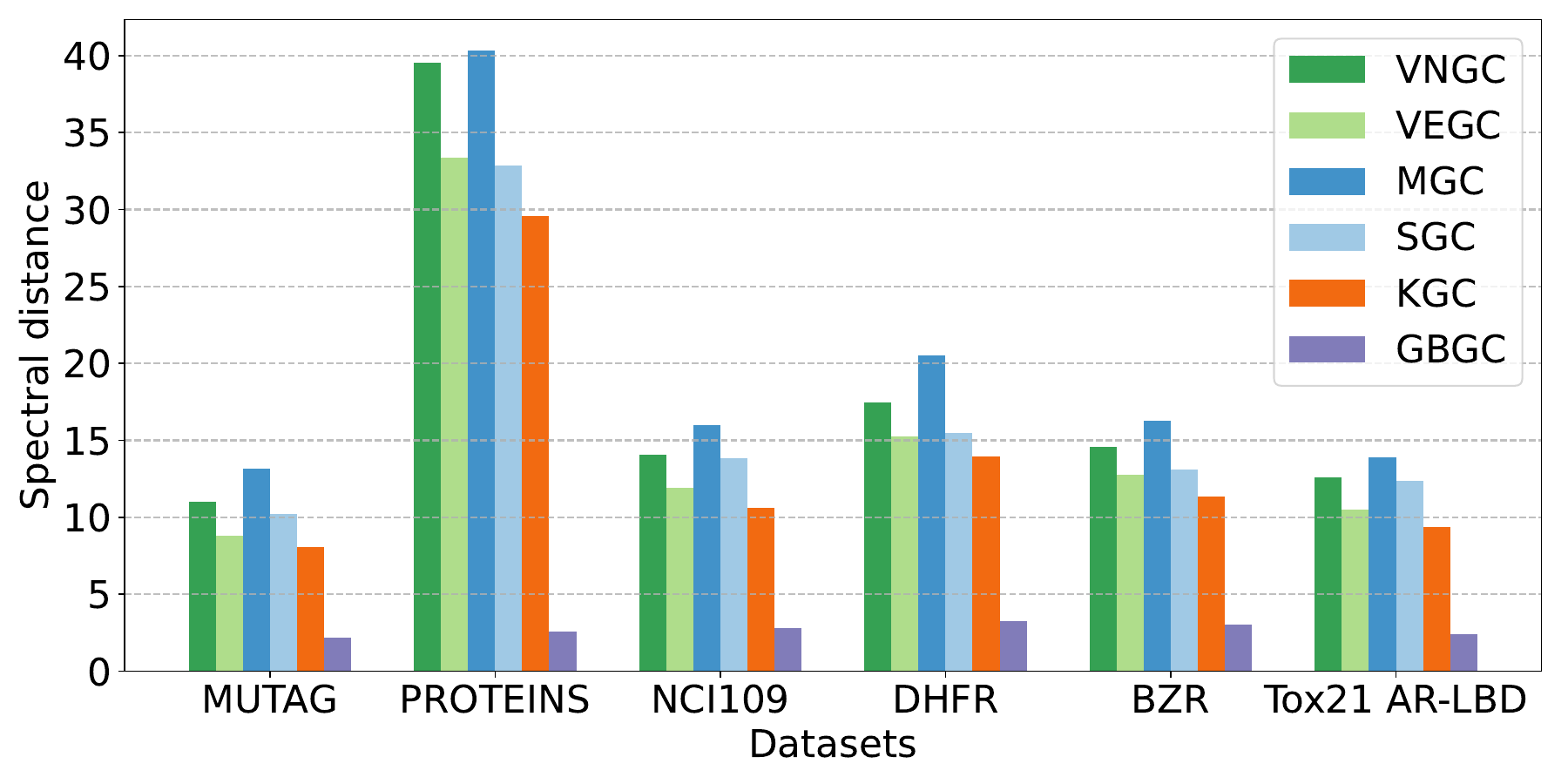}
    \caption{Comparison diagram of $SD$.}
    \label{SD}
\end{figure}

We refer to references \cite{loukas2019graph} and \cite{wilson2008study} and use $SD$ to evaluate the performance of the coarsening methods. Figure \ref{SD} shows an intuitive comparison of $SD$. It can be seen that the spectral distance of the GBGC is better than that of comparison methods. 

\begin{figure}[tb!]
    \centering
    \includegraphics[width=0.95\linewidth]{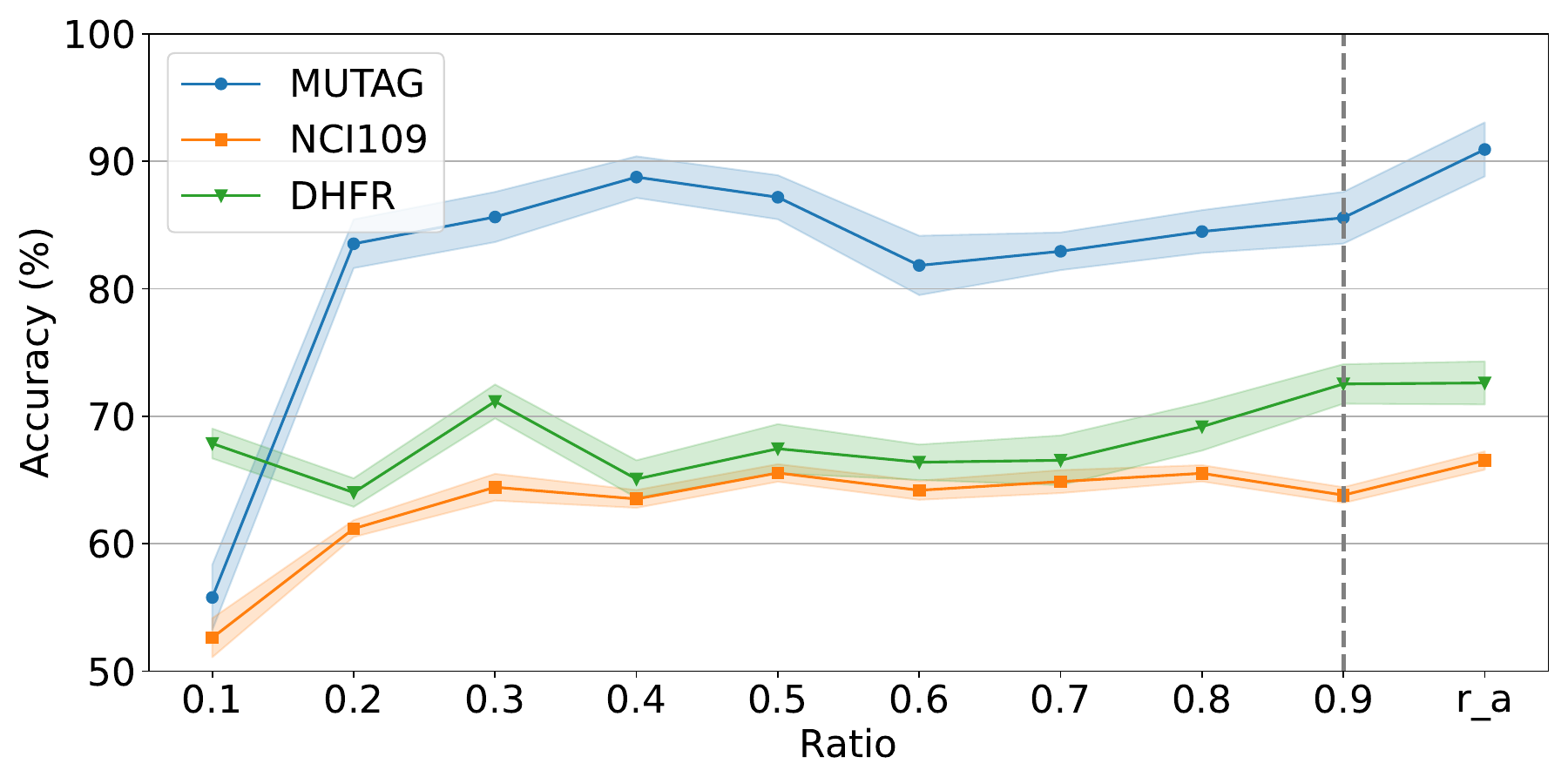}
    \caption{Comparison of different coarsening ratios and adaptive coarsening ratio ($r_a$) of GBGC.}
    \label{linechart}
\end{figure}

\subsection{How does GBGC perform with respect to time consume?\textbf{(RQ2)}}

This section evaluates the efficiency of GBGC and other coarsening methods. Since DosCond’s runtime depends on GNN training, we exclude it from the comparison for fairness and focus on the runtime of GBGC and the other five methods. The runtime comparison is shown in Table \ref{timetable}. From the results in Table \ref{timetable}, we observe that:

\begin{itemize}
\item \textbf{The running time of GBGC is tens to hundreds of times faster than comparison methods, which is significantly lower than other methods.} Especially on the P388H dataset, the advantages of GBGC are even more significant.

\item \textbf{The high efficiency of GBGC can be attributed to its unique granular-ball generation and splitting strategies.} These strategies use global computation for coarse-grained granular-ball initialization and local computation for fine-grained binary splitting, ensuring low time complexity and high efficiency. Importantly, GBGC maintains accuracy while achieving efficiency. By using granular-balls of varying sizes, GBGC effectively captures the original data, often outperforming the FULL in classification accuracy.
\end{itemize}

\begin{figure*}[tb!]
    \centering  
    \subfigure[MUTAG]{
    \label{Fig.sub.1}
    \includegraphics[width=4.5cm,height=3.0cm]{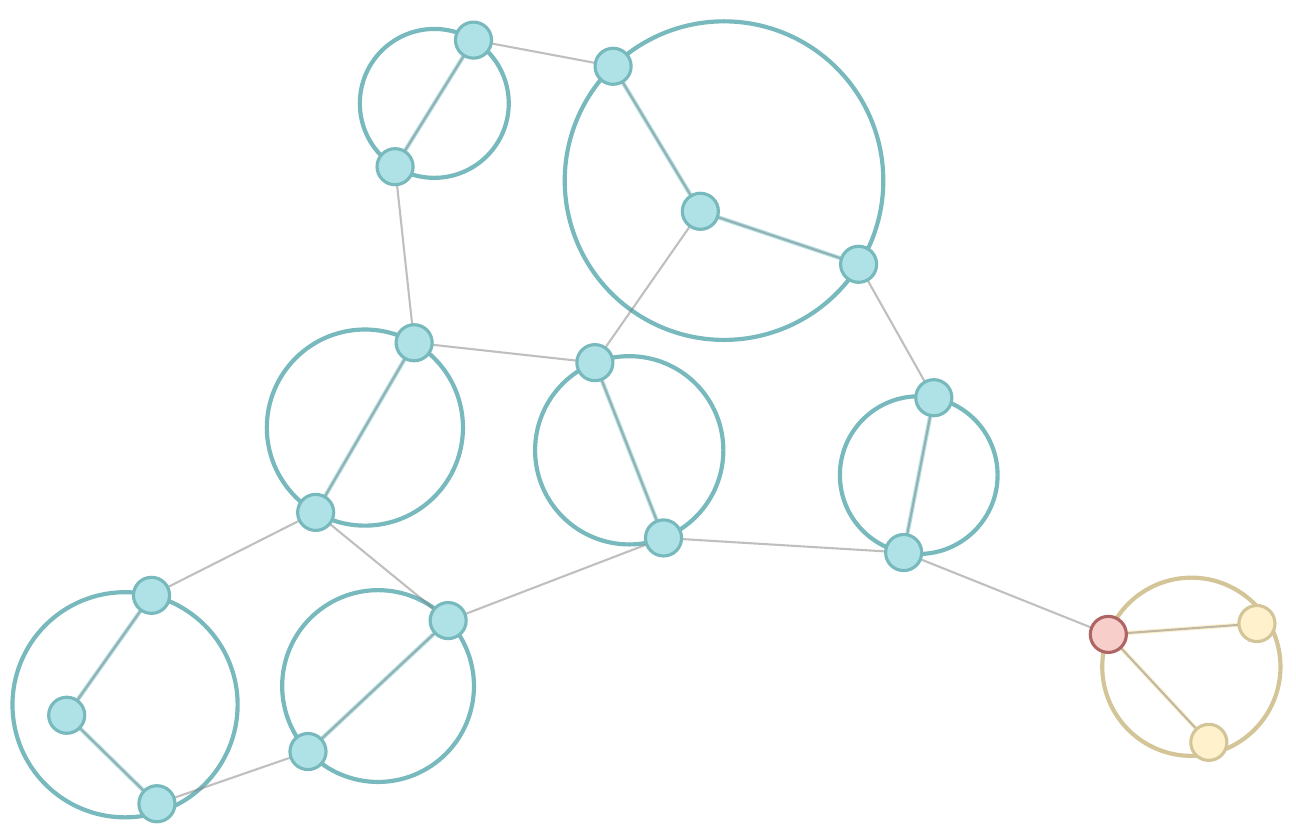}}
    \hspace{2cm} 
    \subfigure[DHFR]{
    \label{Fig.sub.2}
    \includegraphics[width=3.0cm,height=3.0cm]{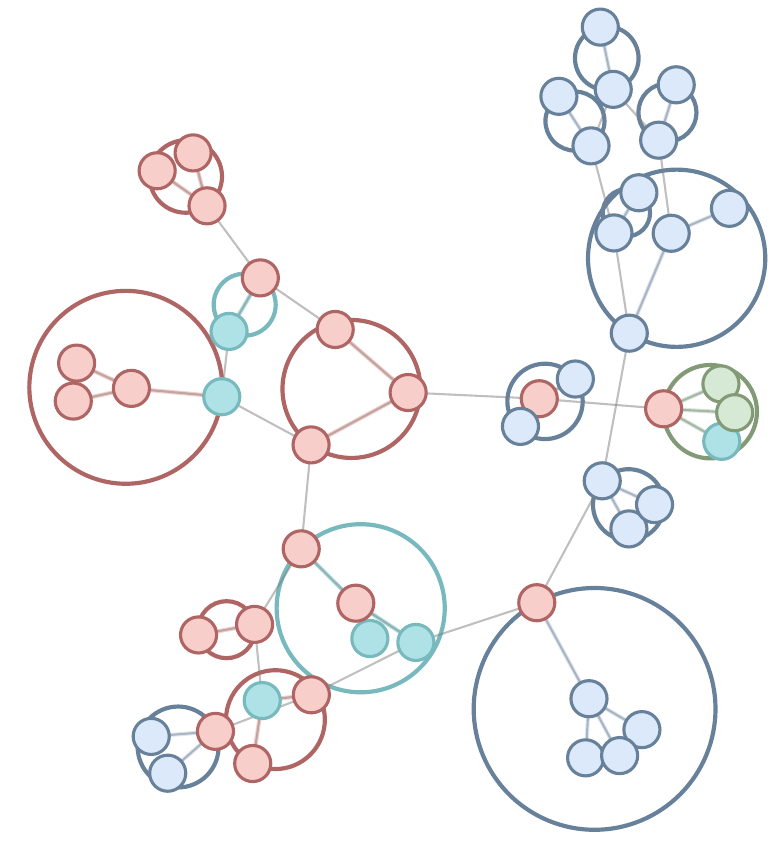}}
    \hspace{2cm} 
    \subfigure[OVCAR-8H]{
    \label{Fig.sub.3}
    \includegraphics[width=3.5cm,height=3.0cm]{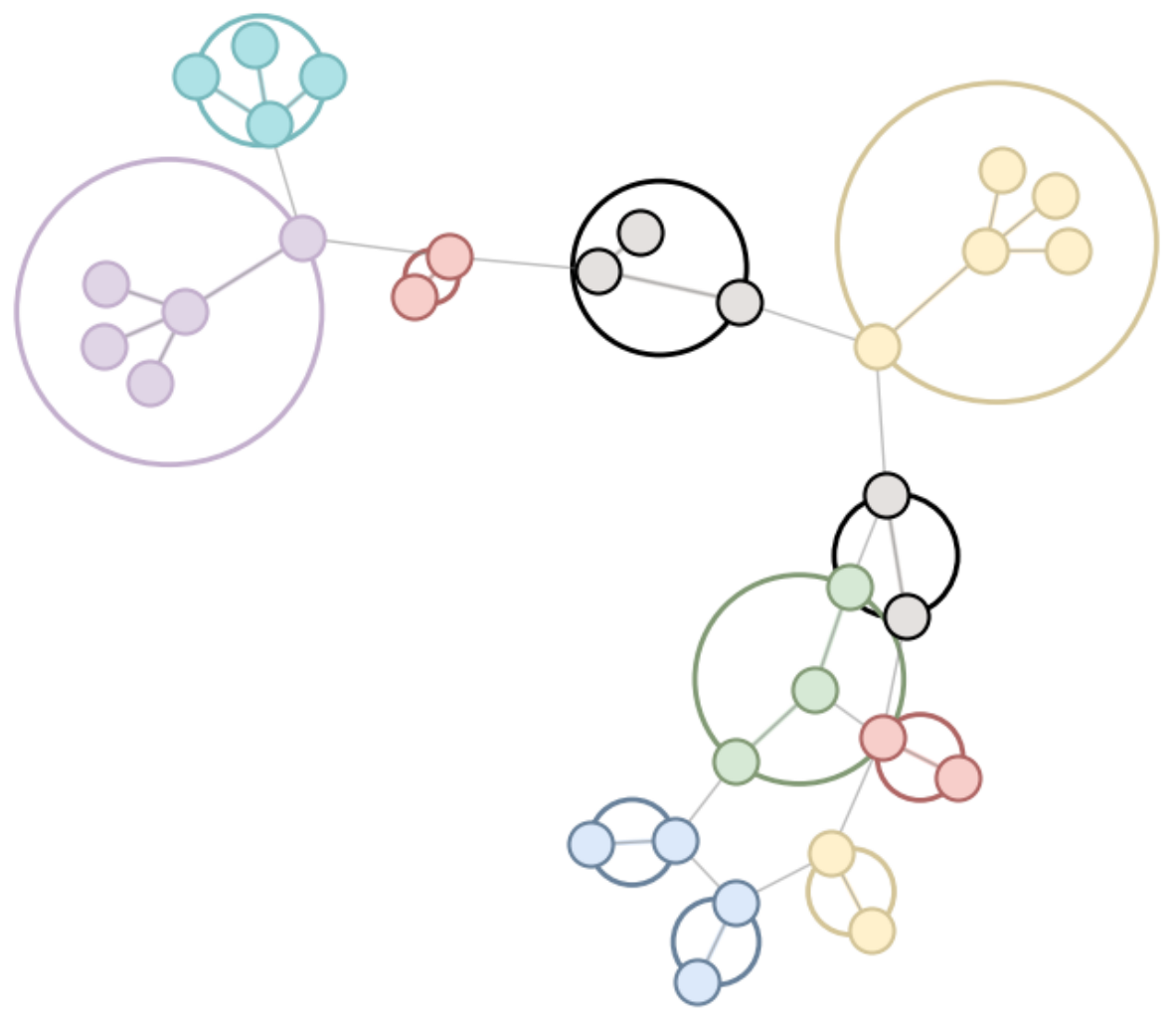}}
    \caption{Visualization of GBGC in different datasets.}
    \label{case}
\end{figure*}

\subsection{Ablation studies \textbf{(RQ3)}}
In this section, we perform ablation experiments on several datasets to examine GBGC, including \textbf{GBGC-w/o fine-grained binary splitting} and \textbf{GBGC-w/o coarse-grained initialization}. The results are shown in Table \ref{ablationtable}. We have the following observations:

\begin{itemize}
\item Removing the fine-grained binary splitting mechanism in GBGC significantly reduces the model's performance, emphasizing its importance. This mechanism is crucial for capturing the graph's inherent granularity and preserving node relationships. Without this mechanism, the model struggles to effectively represent the multi-granularity features of the graph, underscoring its necessity for achieving optimal performance in tasks that rely on precise graph coarsening and representation.

\item Removing the coarse-grained initialization from GBGC leads to a significant performance drop, highlighting its crucial role in the model. It provides an initial representation that captures the graph's broader structure, enabling the model to handle higher-level node relationships. Its absence highlights the substantial contribution it makes to the model’s ability to process and represent the graph effectively, making it an indispensable component to achieve strong performance.
\end{itemize}

\begin{table}[htbp]
    \centering
    \begin{scriptsize}
    \resizebox{\columnwidth}{!}{%
    \begin{tabular}{lllllll}
        \hline
        Datasets & MUTAG & PROTEINS & NCI109 & DHFR  & DD & Avg \\
        \hline
        GBGC      &  \textbf{90.94} & \textbf{65.34} &  \textbf{66.53}   &  \textbf{72.62}  &  \textbf{69.44} &  \textbf{72.97} \\
        -w/o binary splitting  & 89.36  & 63.42 & 63.44 & 72.56  & 68.85 &   71.53 \\
        -w/o initialization & 79.21 & 63.90 & 63.55 & 70.09  & 67.66 & 68.88 \\
        \hline
    \end{tabular}}
    \end{scriptsize}
    \caption{Results of the ablation studies.}
    \label{ablationtable}
\end{table}

\subsection{Hyper-parameter sensitivity analysis \textbf{(RQ4)}}
In this section, we design a non-adaptive GBGC (see Algorithm 4 in Appendix for algorithm design) to evaluate the performance of different $r$. Following the experimental settings of \cite{ChenYaoYangChen2023}, we studied the effect of different $ r \in (0,1) $ with a step size of 0.1 on GBGC performance, as shown in Figure \ref{linechart}. The model performs best when GBGC adaptively generates the coarsening rate, $ r_a $. This is because GBGC uses a mix of large granular-balls for simple data distributions and many smaller ones for complex distributions, effectively representing the data and improving accuracy.

\subsection{Case study \textbf{(RQ5)}}
GBGC retains essential graph structure information, enabling classifiers to use it more efficiently for accurate graph classification. This section focuses on the MUTAG, DHFR, and OVCAR-8H datasets. For MUTAG, GBGC identifies molecular structures related to toxicity; for DHFR, it captures correlations between molecular fragments and drug activity; and for OVCAR-8H, it highlights key molecular features influencing drug response, aiding drug screening and personalized treatment prediction. Figure \ref{case} illustrates GBGC’s analysis, where colored circles represent granular-balls (supernodes), showing its ability to capture node correlations.

\section{Conclusion}

In this work, we propose a new multi-granularity, efficient, and adaptive coarsening method via granular-ball(GBGC). The experimental results show that GBGC has higher accuracy compared with several state-of-the-art methods. At the same time, the speed of GBGC is increased by tens to hundreds of times, showing a low time complexity $O(N ^\frac{3}{2}+ E \sqrt{N})$. Additionally, an intriguing observation is that the accuracy of GBGC is almost consistently higher than that achieved on the original datasets, attributable to its robustness and generalizability. While our current approach focuses on graph classification tasks, other downstream tasks can be explored in the future to further broaden the applicability of our coarsening approach.



\section*{Acknowledgments}
This work was supported in part by the National Natural Science Foundation of China under Grant Nos. 62221005, 62450043, 62222601, and 62176033.



\bibliographystyle{named}
\bibliography{ref}

\end{document}